\documentclass[10pt]{article}

\usepackage{amsthm}
\usepackage{xcolor}

\usepackage{graphicx}
\graphicspath{{/}{fig/}}

\usepackage{enumerate}

\usepackage{float}

\usepackage{hyperref}

\newlength\figureheight
\newlength\figurewidth
\title{
    Towards Large-Scale Scalable MAV Swarms with ROS2 and UWB-based Situated Communication
}

\author{
    Jorge Peña Queralta, Yu Xianjia, \\Li Qingqing, Tomi Westerlund\\[1em]
    \href{https://tiers.utu.fi}{Turku Intelligent Embedded and Robotic Systems (TIERS) Lab}\\{University of Turku, Finland}\\[+4pt]
    Emails: \{jopequ, xianjia.yu, qingqli, tovewe\}@utu.fi
}
\date{}

\begin{document}

\maketitle


\section*{Abstract}

The design and development of swarms of micro-aerial vehicles (MAVs) has recently gained significant traction~\cite{saska2015mav, xu2020decentralized, queralta2020vio}. Collaborative aerial swarms have potential applications in areas as diverse as surveillance and monitoring~\cite{scherer2016persistent, gu2018multiple}, inventory management~\cite{macoir2019uwb, kwon2019robust}, search and rescue~\cite{queralta2020collaborative}, or in the entertainment industry~\cite{waibel2017drone, shule2020uwb}. 

\begin{figure}[b!]
    \centering
    \includegraphics[width=\textwidth]{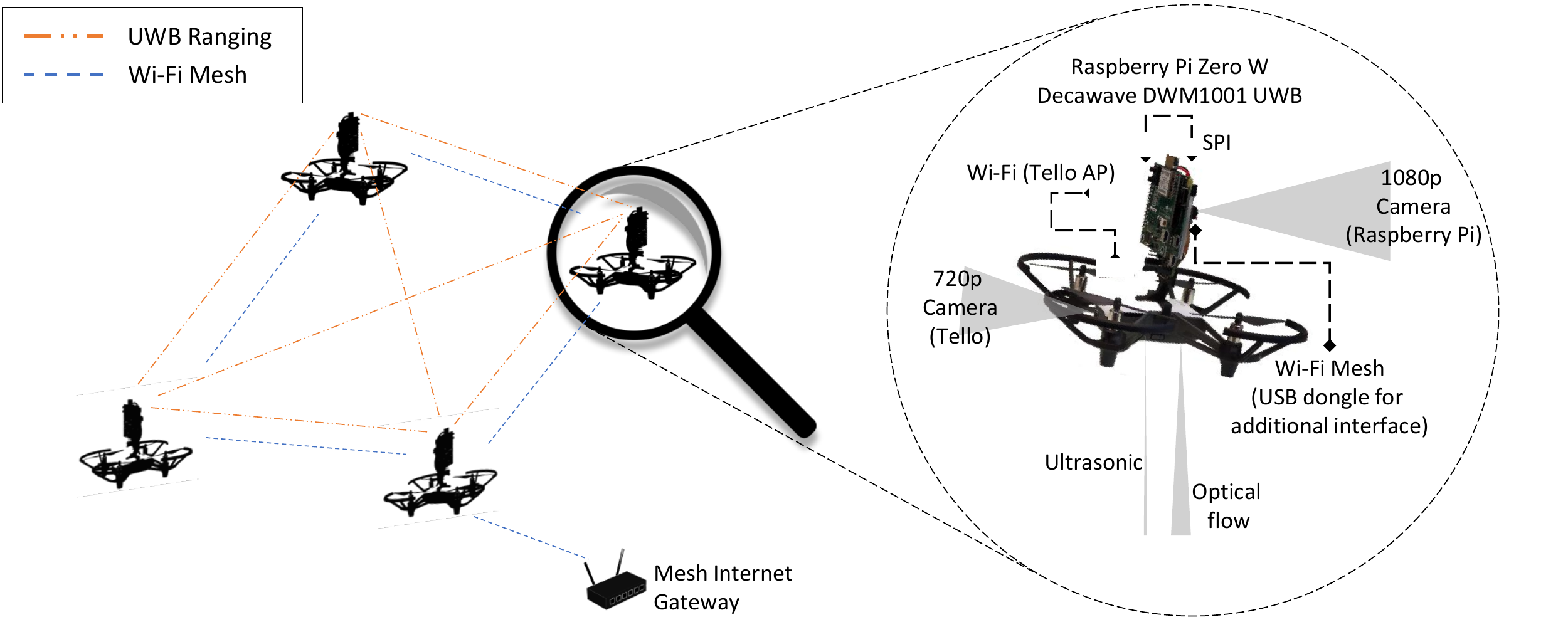}
    \small{\caption{Scalable swarm concept with mesh connectivity and UWB-based relative localization and communication (orientative links). The UWB range can be affected, e.g., by NLOS propagation while Wi-Fi is more prone to interference.}}
    \label{fig:swarm_concept}
\end{figure}

Swarm intelligence has, by definition, a distributed nature. Yet performing experiments in \textit{truly} distributed systems is not always possible, as much of the underlying ecosystem employed 
requires some sort of central control. 
Indeed, in experimental proofs of concept, most research relies on more traditional connectivity solutions (e.g., Wi-Fi)~\cite{campion2018uav}, and robotics middlewares~\cite{quigley2009ros, pinciroli2016buzz}. 
In terms of positioning, external localization solutions, such as motion capture (MOCAP) systems, visual markers, or ultra-wideband (UWB) anchors~\cite{queralta2020uwb, nguyen2016ultra, almansa2020autocalibration} are often used. Alternatively, intra-swarm solutions are often limited in terms of, e.g., range or field-of-view~\cite{walter2019uvdar, queralta2019communication, schilling2018learning}.

Essential to efficient collaboration within aerial swarms are reliable communication and accurate localization. 
Multiple works rely on situated communication as the basis for effective collaboration~\cite{rodrigues2015beyond, mccord2019distributed,  li2019decentralized, queralta2019distributed}. Research and development has been supported by platforms such as the e-puck~\cite{mondada2009puck, millard2017pi}, the kilobot~\cite{rubenstein2014kilobot}, or the crazyflie quadrotors~\cite{giernacki2017crazyflie}. In terms of reproducibility, it is also worth mentioning the MRS Multi-UAV platform~\cite{baca2020mrs}. However, we believe there is a need for inexpensive platforms such as the Crazyflie with more advanced onboard processing capabilities and sensors, while offering scalability and robust communication and localization solutions. In the following, we present a platform for research and development in aerial swarms currently under development, where we leverage Wi-Fi mesh connectivity and the distributed ROS2 middleware together with UWB ranging and communication for situated communication.

\subsection*{Mesh Networking, ROS2 and UWB Localization}

We present a platform for building towards large-scale swarms of autonomous MAVs leveraging the ROS2 middleware, Wi-Fi mesh connectivity, and UWB ranging and communication. The platform is based on the Ryze Tello Drone, a Raspberry Pi Zero W as a companion computer together with a camera module, and a Decawave DWM1001 UWB module for ranging and basic communication. A conceptual illustration is shown in Figure~1. Additionally, a TFMini micro-lidar module can be added to increase the range and accuracy of altitude estimation. When the swarm is deployed, the following provides the underlying methods and systems allowing for complex collaborative behavior:
\begin{enumerate}[(i)]
    \item A Wi-Fi mesh network with predefined settings is deployed using batman-adv~\cite{seither2011routing}, and defining network interfaces for the auto-discovery services in DDS under ROS2 to automatically detect new nodes. A gateway in the mesh network may provide internet access to the swarm.
    \item One-to-one UWB ranging, or alternatively more scalable ranging systems such as SnapLoc~\cite{grobetawindhager2019snaploc}, are utilized based on the node IDs identified within the Wi-Fi mesh. Relative localization is then propagated through the swarm with the possibility for global localization if the position of the gateway or a subset of nodes is known. We assume a common orientation reference, but alternatively a decentralized approach with sliding time window optimization can be applied~\cite{xu2020decentralized}.
    \item Concurrent localization and communication are implemented by embedding data, published to a predefined ROS2 topic, within the UWB ranging messages. We limit this to basic signaling to ensure scalability.
    \item Larger bandwidth is available through the Wi-Fi mesh network while throttling when necessary different ROS2 topics to control network load. 
\end{enumerate}

In conclusion, we have presented an inexpensive MAV platform based on off-the-shelf components opening the door to wider adoption and reproducibility, and more complex proofs of concept in aerial swarms. At the same time, such a multi-MAV system maintains a distributed and scalable nature. Potential applications range from formation control or drone shows to distributed perception and multi-agent search or monitoring.


\bibliographystyle{unsrt}
\bibliography{bibliography}

\end{document}